\documentclass[english]{ccdconf}
\usepackage{booktabs} % For formal tables
\usepackage{multirow}
\usepackage{subfigure}
\usepackage{fancyhdr}

\begin{document}

\title{Self-attention Multi-view Representation Learning with Diversity-promoting Complementarity}
\author{Jian-wei Liu\aref{amss}, Xi-hao Ding\aref{amss},
      Run-kun Lu\aref{amss}, Xionglin Luo\aref{amss}}

\affiliation[amss]{Department of automation, School of Information Science and Engineering,
China University of Petroleum, Beijing, 102249
        \email{{liujw,luoxl}@cup.edu.cn}, 964465871@qq.com, zsylrk@gmail.com}

\maketitle

\begin{abstract}
Multi-view learning attempts to generate a model with a better performance by exploiting the consensus and/or complementarity among multi-view data. However, in terms of complementarity, most existing approaches only can find representations with single complementarity rather than complementary information with diversity. In this paper, to utilize both complementarity and consistency simultaneously, give free rein to the potential of deep learning in grasping diversity-promoting complementarity for multi-view representation learning, we propose a novel supervised multi-view representation learning algorithm, called Self-Attention Multi-View network with Diversity-Promoting Complementarity (SAMVDPC), which exploits the consistency by a group of encoders, uses self-attention to find complementary information entailing diversity. Extensive experiments conducted on eight real-world datasets have demonstrated the effectiveness of our proposed method, and show its superiority over several baseline methods, which only consider single complementary information.
\end{abstract}


\section{INTRODUCTION}

Aiming to make good use of the information from multi-view data and improve the generalization performance, multi-view learning algorithms have made great progress in different tasks, such as classification, regression, and clustering, by utilizing conventional machine learning or deep learning to fully considering the relationships among multiple views \cite{bib1, bib2, bib3, bib4}. And recently, \cite{bib5} analyzes these various algorithms, comes to the conclusion that there are two fundamental assumptions ensuring their success: consistency and complementarity principles. The consistency assumption suggests there is consistent information shared by all views, while the complementarity assumption states each view of multi-data may contain some knowledge that other views do not have. Based on these two assumptions, we review the literature of multi-view learning in recent years, and observe that there are still two drawbacks in many state-of-the-art multi-view learning algorithms.

First, at present, multi-view algorithms can be generally categorized into two types: the first category aims to exploit the consistency, the second one aims to leverage the complementarity among multiple views, and each category only focuses on consensus or complementarity. In detail, the first category usually tries to extract the common latent representation on which all views have minimum disagreement, such as canonical correlation analysis (CCA) class algorithms \cite{bib6, bib7, bib8, bib9, bib10}, which  project two or more views into latent subspaces by maximizing the correlations among projected views, matrix factorization based methods \cite{bib11, bib12, bib13}, which jointly factorize multi-view data into one common centroid representation by minimizing the overall reconstruction loss of different views. And the second category is to explicitly preserve complementary information of different views, such as co-training style algorithms \cite{bib14, bib15, bib16, bib17}, which iteratively train two classifiers on two different views, and each classifier generates its complementary information to help the other classifier to train in the next iteration.

However, both consistency and complementarity of multi-views data are meaningful, the neglect of each aspect will result in the loss of valuable information. In order to address this drawback, multi-view algorithms recently began to develop the third category algorithm, which exploits the consistency and complementarity, simultaneously, such as matrix factorization based methods \cite{bib18, bib19}, which find latent representations composed of common latent factors shared by multiple views and the specific latent factor of each view. But \cite{bib18, bib19} also inherit the shortcomings of matrix factorization, such as they only learn a linear map relationships, can't reflect the non-linear relationship in the multi-view dataset, and require feed all data in one time, lack the ability of dealing with large scale data.

Second, in terms of complementarity, most of existing multi-view learning algorithms only can find representations with single complementarity rather than complementary information with diversity. Srivastava and Salakhutdinov \cite{bib20} propose a deep multi-modal RBM to capture the joint distribution over image and text inputs. Ngiam et al. \cite{bib21} concatenate the final hidden coding of audio and video modalities as input, then map these inputs to a shared representation layer. Cho et al. \cite{bib22} directly input multi-view sequence into RNN encoder to integrate the complementarity of multi-view data. Su et al. \cite{bib23} introduce a multi-view CNN architecture that integrates complementarity among multiple 2D views of an object into a single and compact representation by a view-pooling layer, which performs element-wise maximum operation across the views. In general, all the above algorithms focus on one type of complementarity among multiple views, and they don't consider mining the complementary information with diversity.

To fight against the above mentioned serious deficiencies, in this paper, we propose a new multi-view learning paradigm based on self-attention network, called Self-Attention Multi-View network with Diversity-Promoting Complementarity (SAMVDPC). Specially, SAMVDPC first encodes each view's data into a fixed-length vector representation to exploit the consistency, and then explores complementary information entailing diversity with multiple combination forms by self-attention mechanism, finally concatenates all complementary information into a vector representation, which further be used to make prediction. We may illustrate this idea using an example from a face recognition problem with two views. Given a group of people, we have collected the face information for each person to form two-view dataset. To make classification, first, by building a unique encoder for each view, SAMVDPC encodes each view's data into a fixed-length vector representation and outputs $\mathbf{H}=\left[ {{h}_{1}};{{h}_{2}} \right]\in {{R}^{2\times H}}$. Second, SAMVDPC inputs $\mathbf{H}$ to self-attention mechanism to produce weight matrix $\mathbf{W}=\left[ {{w}_{1}};{{w}_{2}} \right]\in {{}^{2\times 2}}$, then outputs two vectors: ${{w}_{1}}\mathbf{H}$, and ${{w}_{2}}\mathbf{H}$, which can utilize to combine two-view data by different ways for subsequent fusion stage. Finally, SAMVDPC incorporates the concatenated representation $\left[ {{w}_{1}}\mathbf{H},\ {{w}_{2}}\mathbf{H} \right]$ as input and processes these inputs by a forward network to make prediction.

In summary, our contributions we summarize are shown as follows:

(1) We develop a supervised multi-view deep learning algorithm, which utilizes both consistency and complementarity of multiple views, where multiple views' encoders consider the consistency, and self-attention mechanism considers the complementarity;

(2) Compared to \cite{bib18, bib19}, encoders in SAMVDPC can learn nonlinear and hierarchical abstract feature representation for multi-view data, which capture the non-linear relationship and real underlying properties in multi-view dataset.

(3) SAMVDPC can find representations with complementary information possessing diversity rather than single complementarity, and sufficiently reflect the complementarity underlying multi-view data.

(4) We have compared SAMVDPC with other state-of-the-art multi-view learning algorithms and demonstrated its effectiveness, what's more, we also build other baselines deep networks to further analyze SAMVDPC's performance, which explore single complementary by mean-pooling, max-pooling and weighted summation.

\section{Attention mechanism}

In deep neural networks, attention mechanism \cite{bib24} has been developed in the context of encoder-decoder architectures for Neural Machine Translation (NMT) \cite{bib22, bib25}, and rapidly applied to numerous application domains and achieved promising results on several challenging tasks, such as image captioning \cite{bib26}, and summarization \cite{bib27}. Besides, with the development of deep learning, attention mechanism has been free from encoder-decoder contexts, and scholars have gradually developed another kind of attention mechanism, called self-attention mechanism, which highlights the different contributions at different positions of a single sequence in order to compute a context-dependent self-representation of the sequence, and has been used successfully in a variety of tasks including reading comprehension, abstractive summarization, and learning task-independent sentence representations, and yielded impressive performance \cite{bib28, bib29, bib22}.

As researchers deepen their knowledge of attention mechanism, they go a step further and settle down a common structure both for encoder-decoder attention and the self-attention, and formulate attention mechanism as follows:
\begin{equation}
Attention\left( \mathbf{q},\mathbf{V} \right)=softmax\left( f\left( \mathbf{q};\mathbf{\theta } \right) \right)\mathbf{V}.
\end{equation}
In which, $\mathbf{q}$ is a query vector, $\mathbf{V}$ is a set of value vectors. Equation (1) shows that attention mechanism attempts to maps a query and a set of value vectors to an output. The output is computed as a weighted sum of the value vectors, where the weight assigned to each value vector is computed by a function of the query. What's more, when the query $\mathbf{q}$ and value matrix $\mathbf{V}$ are calculated by same inputs, equation (1) represents self-attention mechanism. In contrast, when the query vector and value vectors are calculated by different inputs, equation (1) represents encoder-decoder attention mechanism.

In addition, summarizing the current self-attention mechanism, we find that there are two main ways to implement the self-attention mechanism: 1) the parameters   are directly obtained by the training process, and are independent of model input, such as \cite{bib30}; 2) the parameters   are related to model input, different inputs cause different parameters, such as \cite{bib31}. And in this paper, we adopt the first approach in the way of \cite{bib30} to build a novel supervised multi-view representation learning algorithm with self-attention mechanism, called Self-Attention Multi-View network with Diversity-Promoting Complementarity.

\section{Framework}

\subsection{Notations}

In this paper, bold uppercase characters are used to denote matrices, bold lowercase characters are used to denote vectors, and other characters which are not bold are used to denote scalars. Supposed that $\left\{ {{\mathbf{X}}^{v}}\in {{R}^{{{M}^{v}}\times N}},\mathbf{y}\in {{R}^{N}} \right\}$ is the sample of view \emph{v}, where $v=1,\cdots ,V$, and \emph{V} is the number of views. More specifically, we have \emph{V} views of raw data, and each view can be expressed as $({{\mathbf{X}}^{v}},\mathbf{y})$, where ${{\mathbf{X}}^{v}}=\left[ \mathbf{x}_{1}^{v},\cdots ,\mathbf{x}_{N}^{v} \right]$, $\mathbf{x}_{1}^{v}\in {{R}^{{{M}^{v}}}}$, and $\mathbf{y}=\left[ {{y}_{1}},\cdots ,{{y}_{N}} \right]$.

\subsection{SAMVDPC Architecture}

In this section, we describe the framework of our proposed Self-Attention Multi-View network with Diversity-Promoting Complementarity (SAMVDPC). As shown in Fig. 1(a), the architecture of SAMVDPC is made up of Encoder-Block, self-attention mapping, and fully connected layer, and the detailed self-attention mapping processes are shown in Fig. 1(b). We describe each of the constituents in the following subsections.

\begin{figure*}
\centering
\subfigure[]{
\includegraphics[height=8.46cm,width=10.34cm]{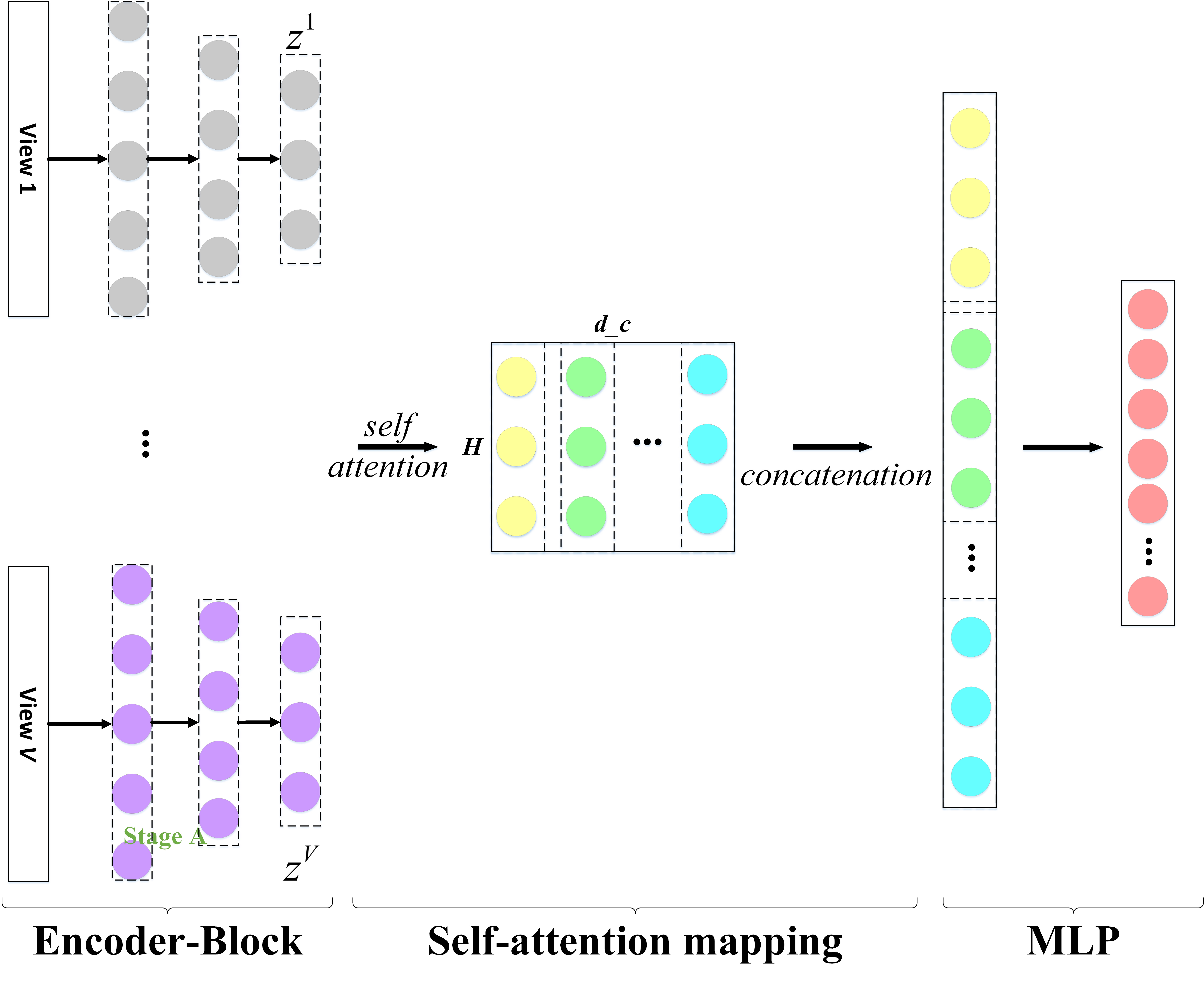}
}
\quad
\subfigure[]{
\includegraphics[height=8.46cm,width=3.97cm]{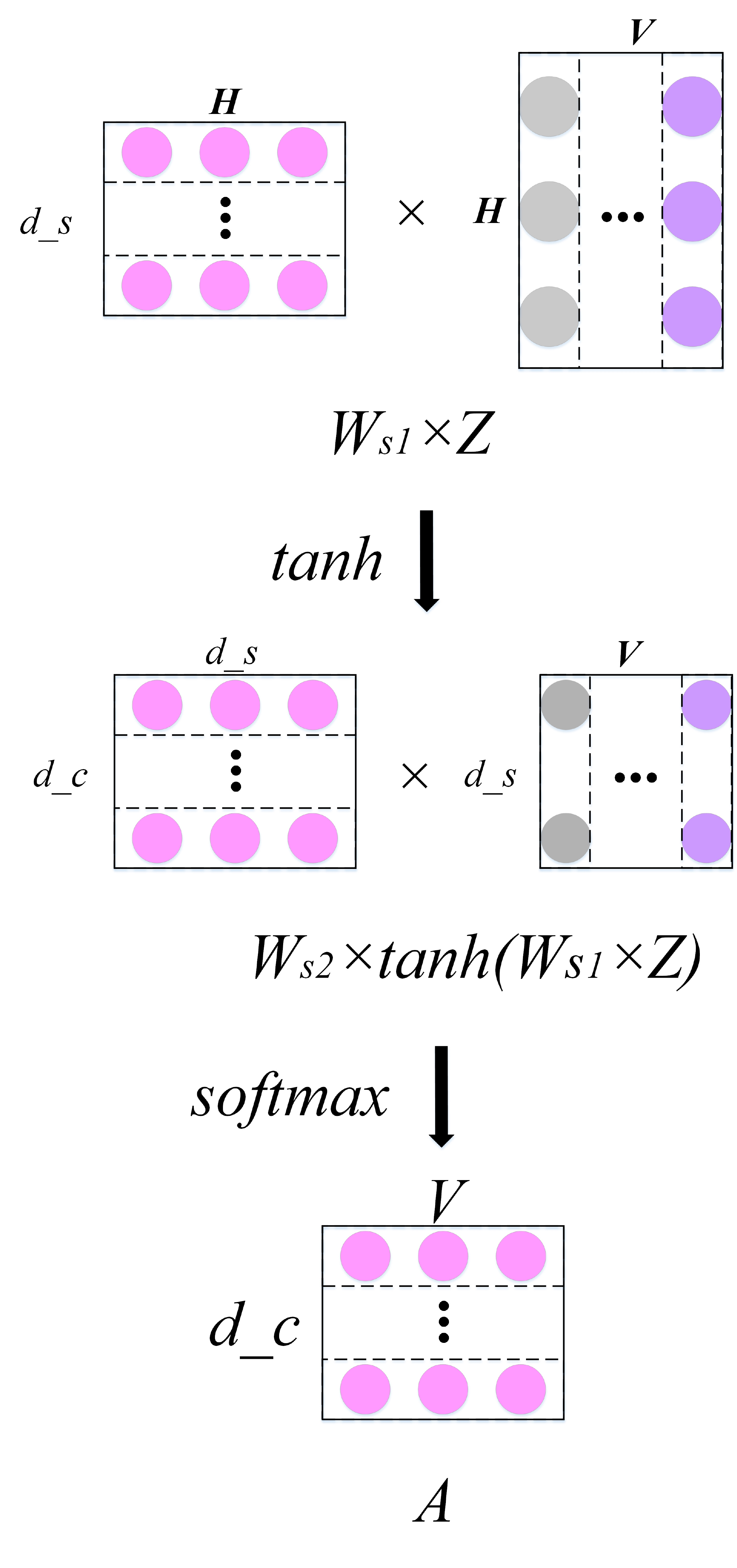}
}
\caption{MVCapsNet Architecture. Fig. 1: (a) is the architecture of SAMVDPC. (b) is concrete self-attention mapping implementation processes.}
\end{figure*}

\textbf{Encoder-Block:} As shown in Fig. 1(a), Encoder-Block is composed with \emph{V} same encoders to extract each view's feature. The initial model parameters of each encoder are initialized by the encoder of a corresponding auto-encoder, which will be explained more detailed in section 4.4. From these encoders, \emph{V} hidden features (${{\mathbf{z}}^{v}}\in {{R}^{H\times 1}}, v=1,\cdots ,V$) can be obtained and they will be stacked horizontally and combined into a feature matrix:
$\mathbf{Z}=\left[ {{\mathbf{z}}^{1}},\cdots {{\mathbf{z}}^{V}} \right],\ {{\mathbf{z}}^{v}}\in {{R}^{H\times 1}}$,
where \emph{H} is the number of dimension of hidden feature vector ${{\mathbf{z}}^{v}}$.

\textbf{Self-attention mapping:} The self-attention mechanism takes the whole hidden states matrix $\mathbf{Z}$ as input, outputs a matrix $\mathbf{A}$, and each row of $\mathbf{A}$ is a vector of weights ${{\mathbf{a}}_{i}}$:
\begin{equation}
\mathbf{A}=\left[ \begin{matrix}
   {{\mathbf{a}}_{1}}  \\
   \vdots   \\
   {{\mathbf{a}}_{d\_c}}  \\
\end{matrix} \right]=softmax\left( {{\mathbf{W}}_{s2}}\tanh \left( {{\mathbf{W}}_{s1}}{{\mathbf{Z}}^{T}} \right) \right),
\end{equation}
where, $\mathbf{A}\in {{R}^{d\_c\times V}}$, ${{\mathbf{W}}_{s1}}\in {{R}^{d\_s\times H}}$, ${{\mathbf{W}}_{s2}}\in {{R}^{d\_c\times d\_s}}$, $d\_s$ is a hyper parameter we can set arbitrarily, and the softmax() is operated along the second dimension of its input. Inspired by [30], Equation (2) also can be deemed as a 2-layer MLP without bias, whose hidden unit numbers are $d\_s$, and parameters are $\left\{ {{\mathbf{W}}_{s2}},{{\mathbf{W}}_{s1}} \right\}$. Finally, we compute the $d\_s$ weighted sums by multiplying $\mathbf{A}$ and $\mathbf{Z}$:
\begin{equation}
\mathbf{M}=\mathbf{A}{{\mathbf{Z}}^T},\ \mathbf{M}\in {{R}^{d\_c\times H}}.
\end{equation}
It is worth noting that each row of $\mathbf{M}$ is a unique nonlinear combinations of multiple views data, and the self-attention mechanism outputs formulate $d\_s$ kinds of nonlinear combination of multiple views data. In our experiment part, the value of $d\_s$ is set to \emph{V}.

\textbf{MLP:} We concatenate each row of   to produce a multi-view representation containing multiple combinations of multi-view data to extract complementary information entailing diversity. Then we input this representation to 2-layer MLP, and make prediction.

\subsection{Objective Function and Regularization}

The embedding matrix $\mathbf{M}$ always suffer from redundancy problems because the self-attention mechanism often provides similar summation weights for all the $d\_s$ hops. Inspired by \cite{DBLP:conf/iclr/LinFSYXZB17}, we also add regularization to encourage the diversity of summation weights vectors across different hops of attention. Thus, in this paper, our objective function is consist of cross entropy loss and regularization, and can be formulated as follows:
\begin{equation}
L=cross\_entropy(y,\hat{y})+\lambda \left\| \mathbf{A}{{\mathbf{A}}^{T}}-\mathbf{I} \right\|_{F}^{2},
\end{equation}
where $\lambda $ is regularization parameters, and \textbf{I} is a unit diagonal matrix.

\section{Experiment}

In this section, we experimentally evaluate SAMVDPC in classification task on eight real world multi-view data sets by comparing it to other baseline algorithms, and design a set of exploratory experiments to validate properties of the self-attention mechanism in SAMVDPC, finally analyse the convergence of our proposed algorithm.

\subsection{Datasets}

In this paper, we use eight real-world multi-view data sets to verify the performance of SAMVDPC, including Leaves, Reuters, YaleFace, BBC, Cornell, Texas, Washington, and Wisconsin datasets. Leaves and YaleFace are two image dataset, Reuters and BBC are two text dataset, Cornell, Texas, Washington, and Wisconsin dataset are four subset of data sets selected from WebKB data sets, and WebKB are webpage dataset. The properties of data sets are summarized in Table 1.
% Please add the following required packages to your document preamble:
% \usepackage{multirow}
\begin{table}
    \caption{Characteristics of the datasets}
    \label{tab:Characteristics}
    \begin{tabular}{ccccc}
    \toprule
    \multirow{2}{*}{\textbf{Data Set}} & \multicolumn{4}{c}{\textbf{Characteristics}}                                                                                                                          \\ \cline{2-5}
                                       & \textbf{\begin{tabular}[c]{@{}c@{}}Instances\\ numbers\end{tabular}} & \textbf{V} & \textbf{K} & \textbf{\begin{tabular}[c]{@{}c@{}}Dimension\\ numbers\end{tabular}} \\ \hline
    Leaves                            & 96                                                                   & 3          & 6          & 64 for all                                                           \\
    Reuters                            & 1200                                                                 & 5          & 6          & 2000 for all                                                         \\
    YaleFace                           & 256                                                                  & 2          & 8          & 2016 for all                                                         \\
    BBC                         & 685                                                                  & 4          & 5          & 4659/4633/4665/4684                                                  \\
    Cornell                            & 195                                                                  & 2          & 5          & 1703/585                                                             \\
    Texas                              & 187                                                                  & 2          & 5          & 1703/561                                                             \\
    Washington                         & 230                                                                  & 2          & 5          & 1703/690                                                             \\
    Wisconsin                          & 265                                                                  & 2          & 8          & 1703/795                                                             \\
    \bottomrule
    \end{tabular}
\end{table}

\subsection{Comparison Algorithms and Baseline Models}

We evaluate the SAMVDPC performance in classification tasks by comparing it with several state-of-the-art multi-view learning algorithms based on matrix factorization, such as GMVNMF \cite{bib32}, multiNMF \cite{bib12}, MVCC \cite{bib13}, DICS \cite{bib18}, and some our designed deep neural network baseline models with three sorts of fusion strategies replacing with our self-attention mechanism, including max-pooling model, mean-pooling model, and weighted summation model. For fair comparison, in terms of matrix factorization algorithms, we choose the parameters within the range that author suggested to obtain good latent representations, and input these representations to KNN($k=1$) for classification; in terms of deep neural network baseline models, we instead of the self-attention mechanism with max-pooling, mean-pooling, or weighted summation fusion, maintain the remaining structure unchanged, and remove the regularization in our objective function.

% Please add the following required packages to your document preamble:
% \usepackage{multirow}
\begin{table*}
    \caption{Hyperparameters on each data set}
    \label{tab:Hyperparameters}
    \begin{center} 
    \begin{tabular}{cccccc}
    \toprule
    \multirow{2}{*}{\textbf{Data Set}} & \multicolumn{3}{c}{\textbf{\begin{tabular}[c]{@{}c@{}} Units number \\ of each encoder layer\end{tabular}}} & \textbf{\begin{tabular}[c]{@{}c@{}}Diversity of\\complementarity\end{tabular}} & \multirow{2}{*}{\textbf{\begin{tabular}[c]{@{}c@{}} Size of \\ mini-batch\end{tabular}}} \\ \cline{2-5}
    & ${{l}_{1}}$             & ${{l}_{2}}$                & ${{l}_{3}}$               & $d\_c$                                                          & \multicolumn{1}{l}{}                                                       \\ \midrule
    Leaves                             & $\quad $64$\quad $            & $\quad $32$\quad $                                  & $\quad $16$\quad $                                 & 3                                                                               & 4                                                                          \\
    Reuters                            & 2048                                & 1024                                & 512                                & 2                                                                               & 16                                                                         \\
    YaleFace                           & 1024                                & 512                                 & 512                                & 2                                                                               & 32                                                                         \\
    BBC                                & 1024                                & 512                                 & 512                                & 4                                                                               & 32                                                                         \\
    Cornell                            & 1024                                & 512                                 & 128                                & 2                                                                               & 16                                                                         \\
    Texas                              & 1024                                & 512                                 & 128                                & 2                                                                               & 16                                                                         \\
    Washington                         & 1024                                & 512                                 & 128                                & 2                                                                               & 16                                                                         \\
    Wisconsin                          & 1024                                & 512                                 & 128                                & 2                                                                               & 16                                                                         \\
    \bottomrule
    \end{tabular}
\end{center}
\end{table*}

% Please add the following required packages to your document preamble:
% \usepackage{multirow}
\begin{table*}
    \caption{Accuracy of different methods}
    \label{tab:acc}
    \begin{tabular}{ccccccccc}
    \hline
    \multirow{2}{*}{\textbf{Method}} & \multicolumn{8}{c}{\textbf{ACC(\%)}}                                                                                                                        \\ \cline{2-9}
                                     & \textbf{Leaves} & \textbf{YaleFace} & \textbf{Reuters}  & \textbf{BBC}      & \textbf{Cornell}  & \textbf{Texas} & \textbf{Washington} & \textbf{Wisconsin} \\ \hline
    \textbf{GNMF}                    & 95.0$\pm $0          & 50.0$\pm $2.5          & 40.8$\pm $1.2          & 38.0$\pm $1.5          & 41.0$\pm $1.8          & 57.9$\pm $1.8       & 69.6$\pm $2.2            & 52.8$\pm $1.4           \\
    \textbf{MultiNMF}                & 95.0$\pm $0          & 64.2$\pm $4.2          & 52.7$\pm $0.2          & 73.1$\pm $0.2          & 49.7$\pm $7.7          & 68.7$\pm $3.4       & 59.3$\pm $2.6            & 50.3$\pm $3.5           \\
    \textbf{MVCC}                    & 100$\pm $           & 33.3$\pm $6.9          & 54.4$\pm $1.9          & \textbf{95.8$\pm $2.6} & 60.8$\pm $5.0          & 64.7$\pm $5.5       & 62.8$\pm $3.8            & 64.3$\pm $2.7           \\
    \textbf{DICS}                    & 97.9$\pm $2.5        & 89.1$\pm $3.2          & 70.3$\pm $4.0          & 90.2$\pm $2.4          & \textbf{72.8$\pm $6.1} & 81.6$\pm $4.0       & \textbf{77.4$\pm $6.0}   & 85.1$\pm $4.5           \\
    \textbf{MAX-Pooling}             & 100$\pm $0           & 90.0$\pm $4.6          & 71.2$\pm $3.4          & 80.5$\pm $7.6          & 71.3$\pm $8.7          & 74.7$\pm $5.2       & 67.5$\pm $8.2            & 86.2$\pm $7.7           \\
    \textbf{MEAN-Pooling}            & 100$\pm $0           & 90.6$\pm $5.3          & 71.2$\pm $4.3          & 83.3$\pm $6.7          & 70.9$\pm $5.5          & 76.6$\pm $4.0       & 70.0$\pm $4.9            & 84.8$\pm $5.2           \\
    \textbf{Weighted Sum}            & 100$\pm $0           & 92.9$\pm $4.8          & \textbf{72.8$\pm $4.7} & 87.2$\pm $4.5          & 72.5$\pm $13           & 76.3$\pm $4.9       & 66.9$\pm $7.8            & \textbf{86.3$\pm $4.3}           \\
    \textbf{AMVDPC}                  & \textbf{100$\pm $0}  & \textbf{94.0$\pm $2.2} & 70.0$\pm $5.2          & 93.5$\pm $2.4          & 72.2$\pm $4.9          & \textbf{85.3$\pm $4.0}       & 75.0$\pm $6.1            & 84.0$\pm $5.2           \\ \hline
    \end{tabular}
\end{table*}

GMVNMF is an NMF-based algorithm by merging local geometrical structure information of each view in a multi-view feature extraction framework. The extracted feature considered the inner-view relatedness between data, and further can be used to complete various tasks. We select parameters ${{\lambda }_{f}}$, $\mu $ to 0.01, and 10 as author suggested, respectively.

MultiNMF is an NMF-based multi-view algorithm, in terms of matrix factorization, it requires coefficient matrices learnt from different views to be softly regularized towards a common consensus matrix, which reflect the information of multi-view data and can be used to make classification. We select the values of regularization parameter $\lambda $ are $10^{-3}$, $10^{-2}$, $10^{-1}$, and 1.

MVCC is a novel multi-view method based on concept factorization with local manifold regularization, which also drives a common consensus representation for multiple views. We set parameter $\alpha $ to 100, and both select the values of parameters $\beta $ and $\gamma $ are 50, 100, 200, 500, and 1000.

DICS is an NMF-based multi-view learning algorithm, by exploring the discriminative and nondiscriminating information existing in common and view-specific parts among different views via joint non-negative matrix factorization, and produce discriminative and non-discriminative feature from all subspaces. And then, discriminative and nondiscriminative features are further used to produce classification results. We select parameters $\alpha $ and $\beta $ within a small range of $\left[ 0,1 \right]$, and set parameter $\gamma $ to 1.

\begin{figure*}[ht]
\begin{center}
\begin{minipage}{0.30\linewidth}
\centerline{\includegraphics[width=1\textwidth]{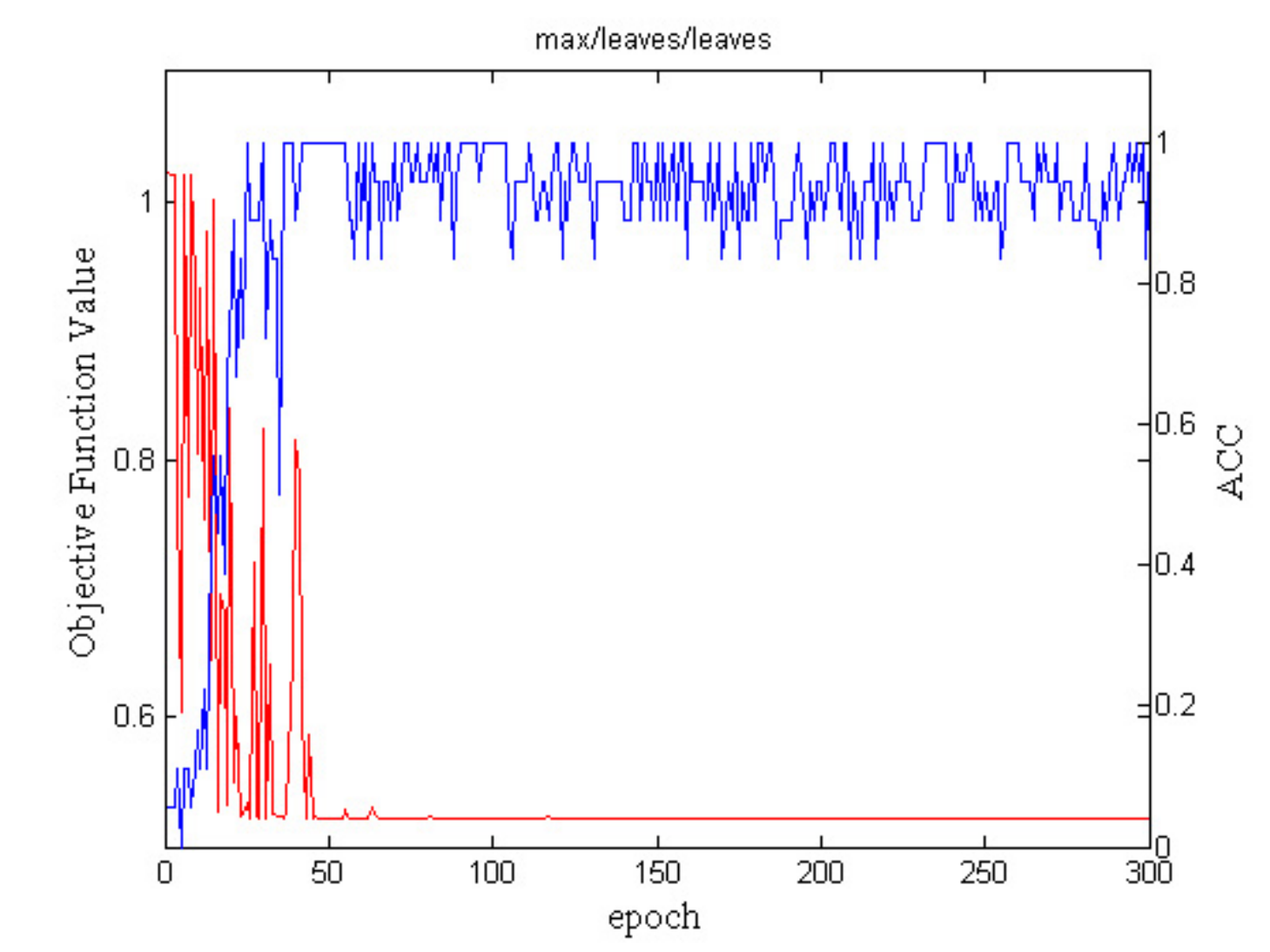}}
\centerline{(a1) leaves-max-pooling}
\end{minipage}
\qquad
\begin{minipage}{0.30\linewidth}
\centerline{\includegraphics[width=1\textwidth]{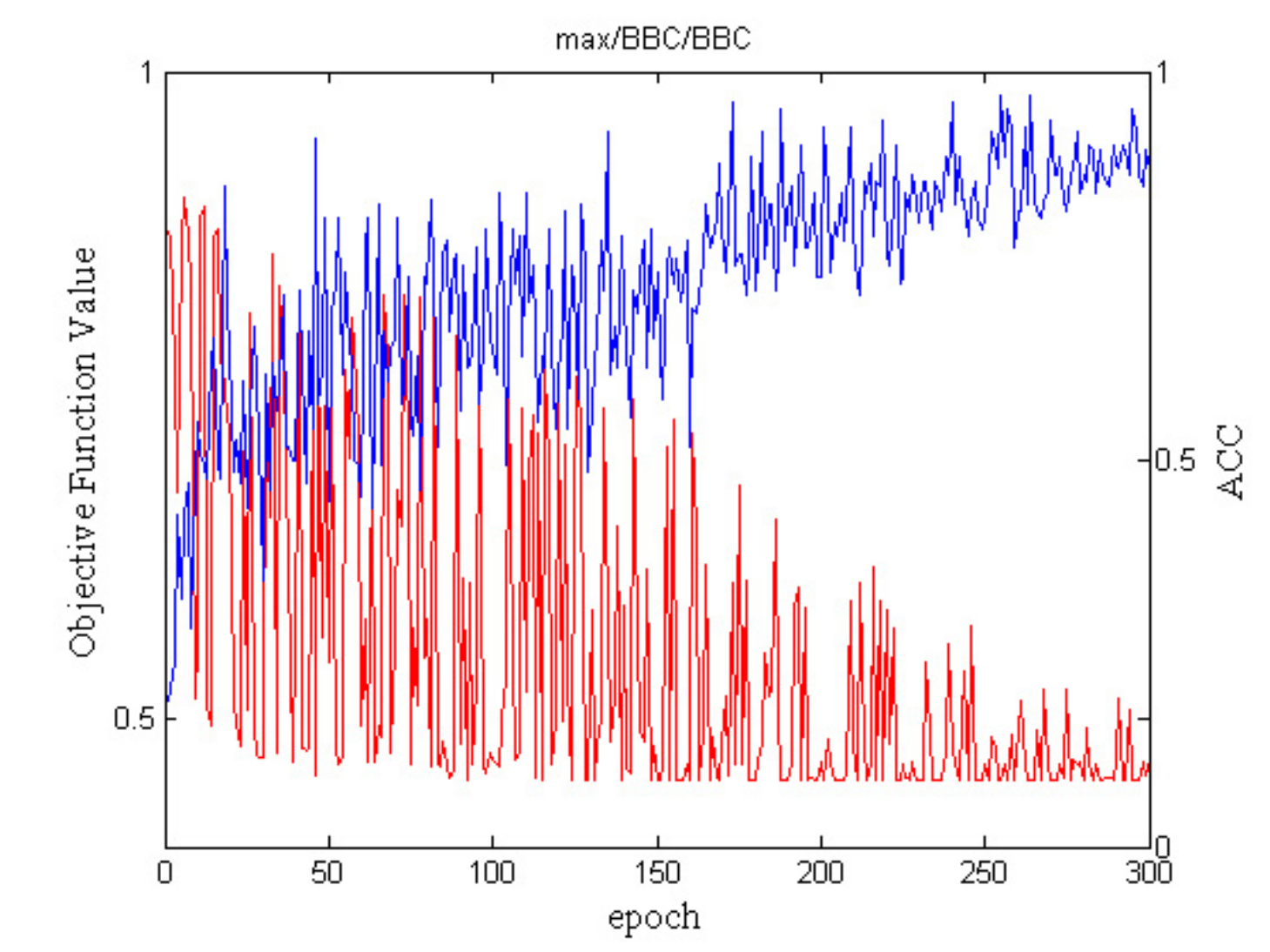}}
\centerline{(b1) BBC-max-pooling}
\end{minipage}
\qquad
\begin{minipage}{0.30\linewidth}
\centerline{\includegraphics[width=1\textwidth]{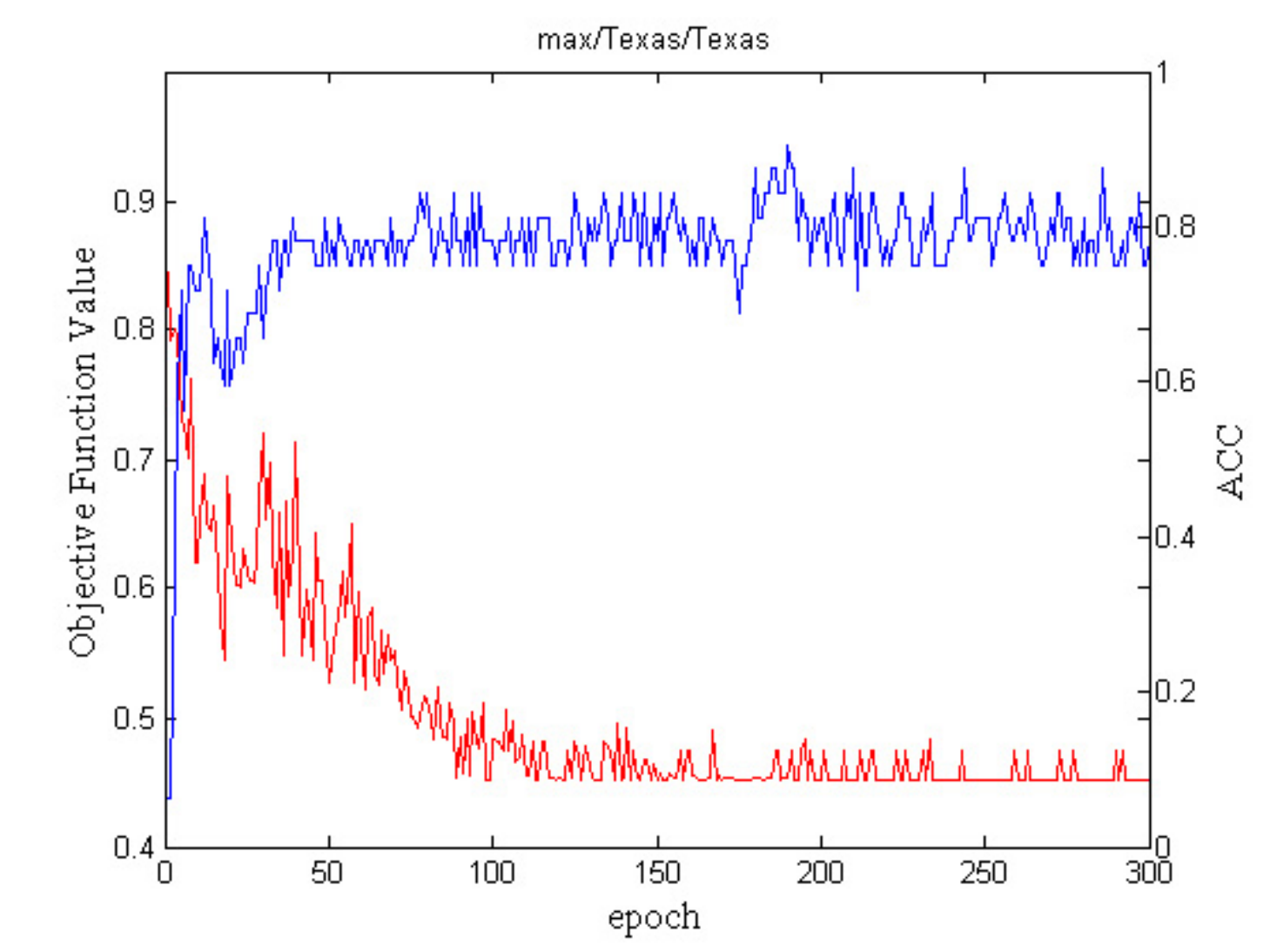}}
\centerline{(c1) Texas-max-pooling}
\end{minipage}

\begin{minipage}{0.30\linewidth}
\centerline{\includegraphics[width=1\textwidth]{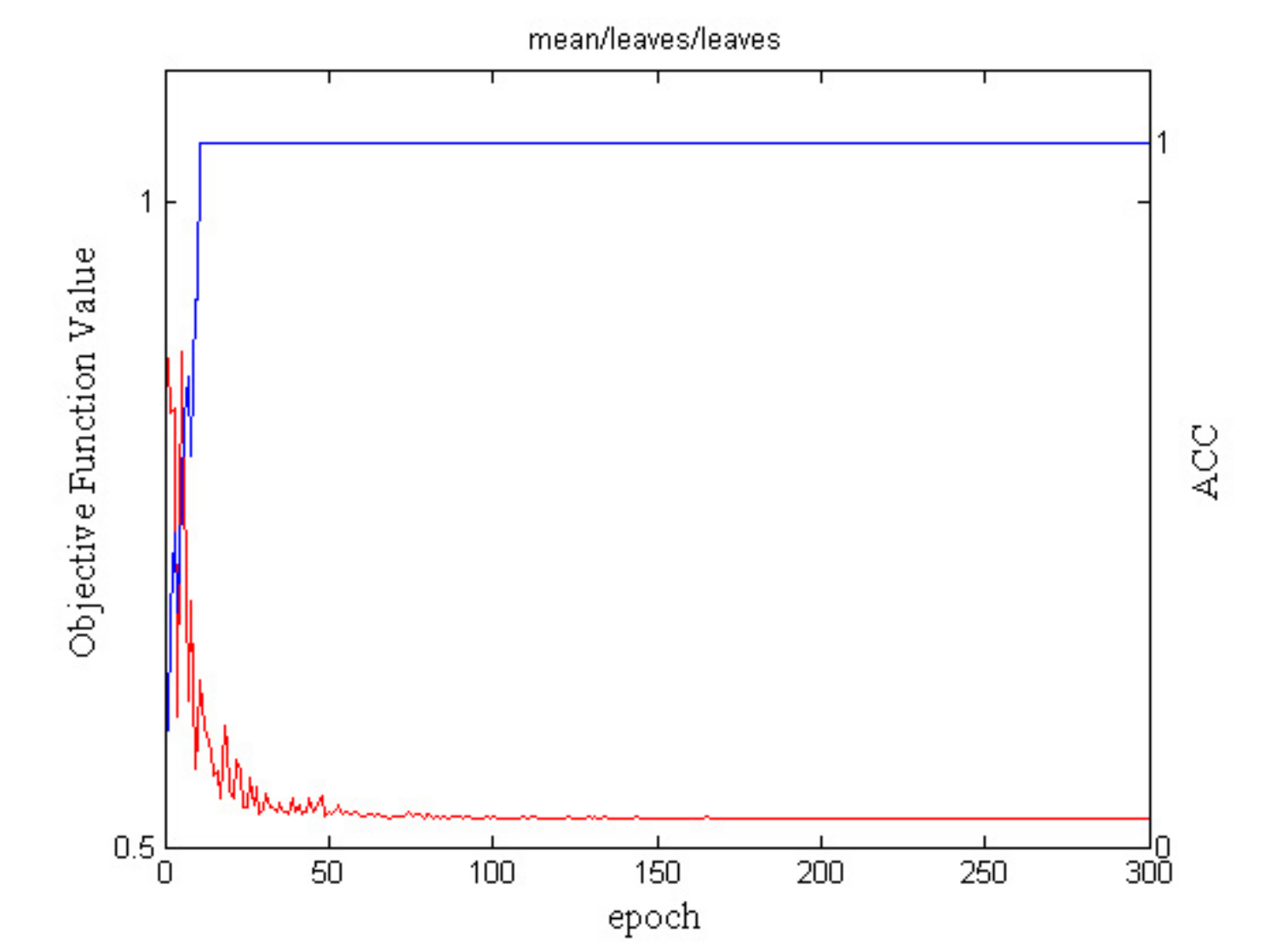}}
\centerline{(a2) leaves-mean-pooling}
\end{minipage}
\qquad
\begin{minipage}{0.30\linewidth}
\centerline{\includegraphics[width=1\textwidth]{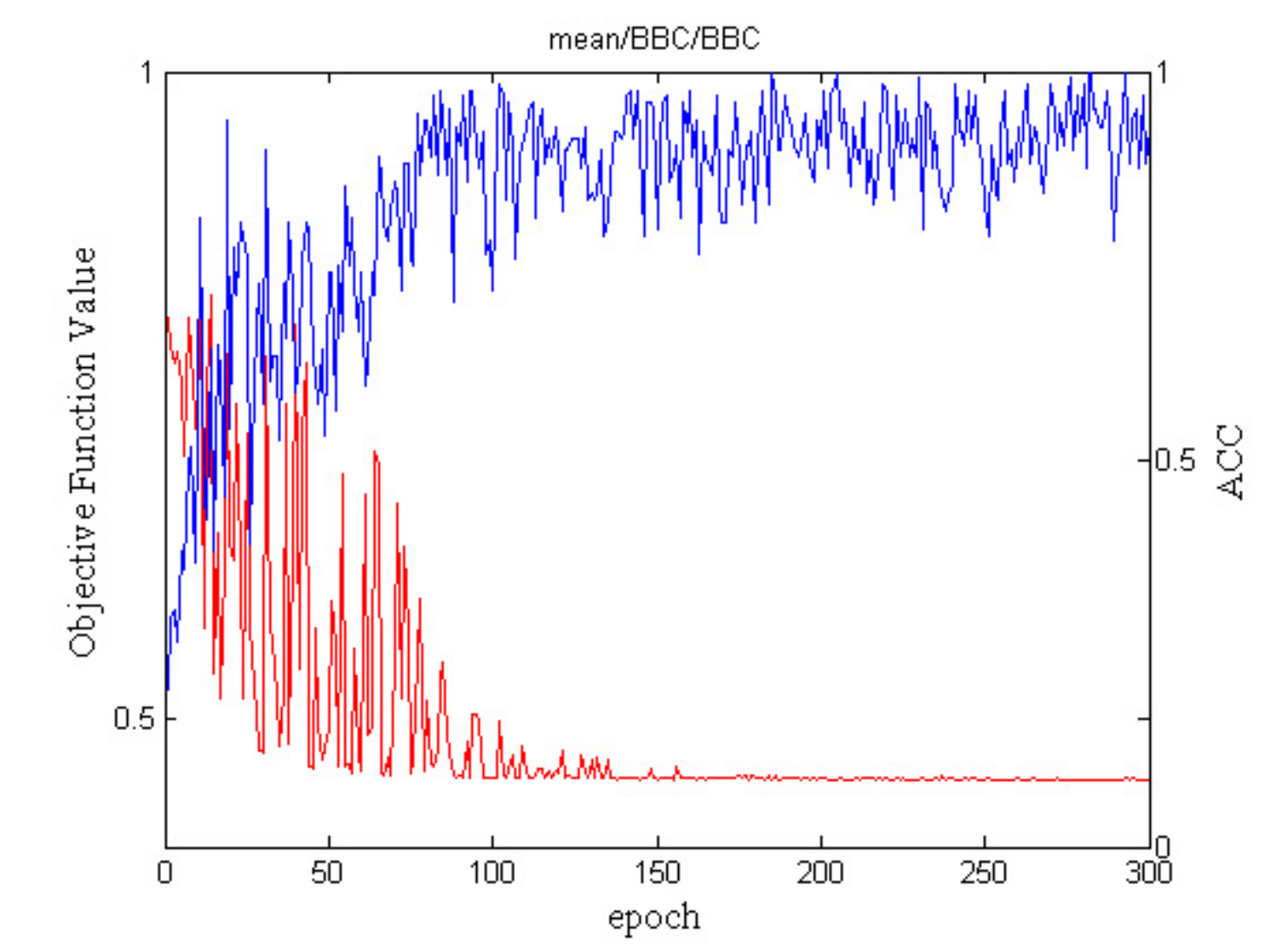}}
\centerline{(b2) BBC-mean-pooling}
\end{minipage}
\qquad
\begin{minipage}{0.30\linewidth}
\centerline{\includegraphics[width=1\textwidth]{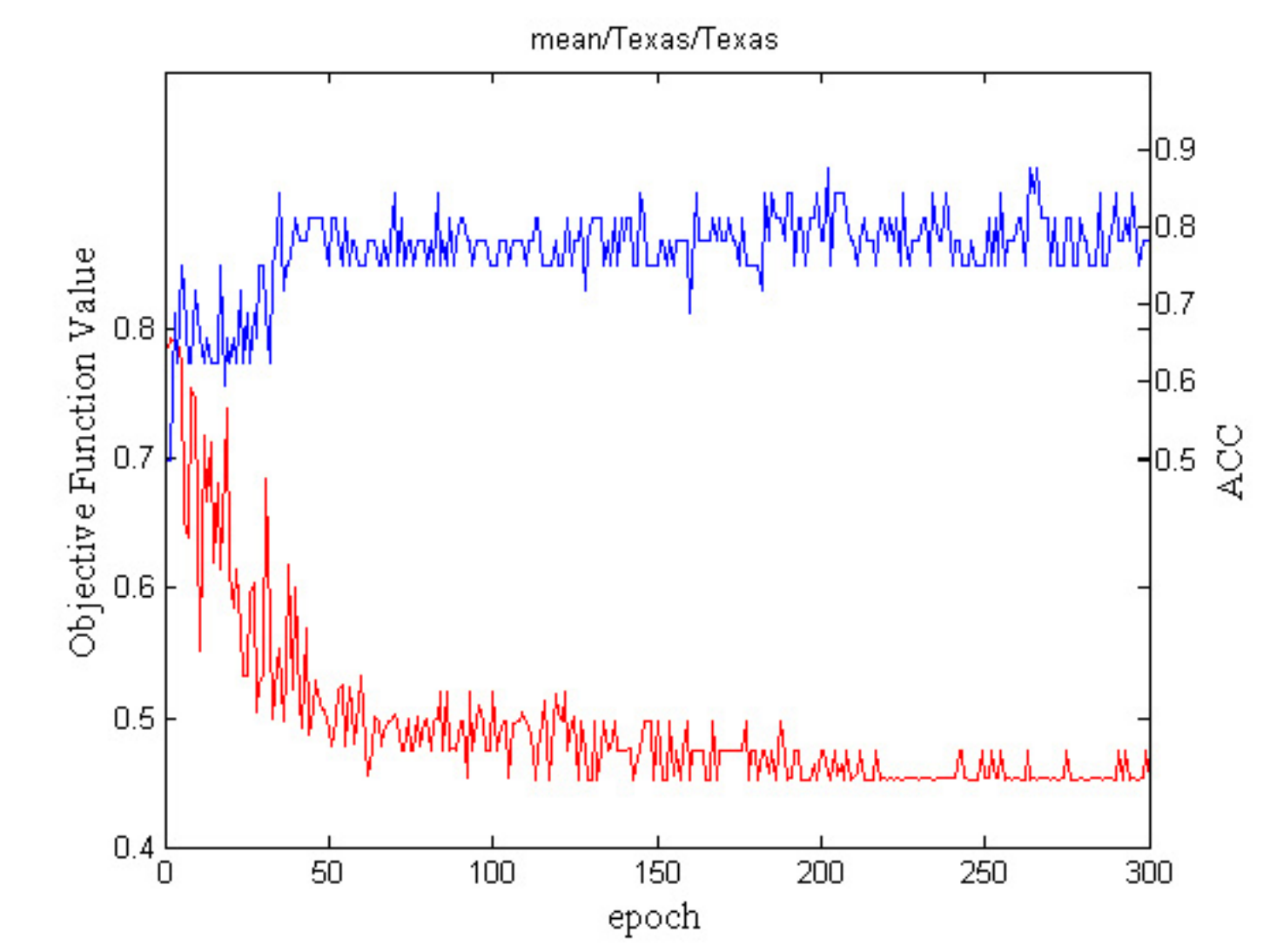}}
\centerline{(c2) Texas-mean-pooling}
\end{minipage}
\qquad

\begin{minipage}{0.30\linewidth}
\centerline{\includegraphics[width=1\textwidth]{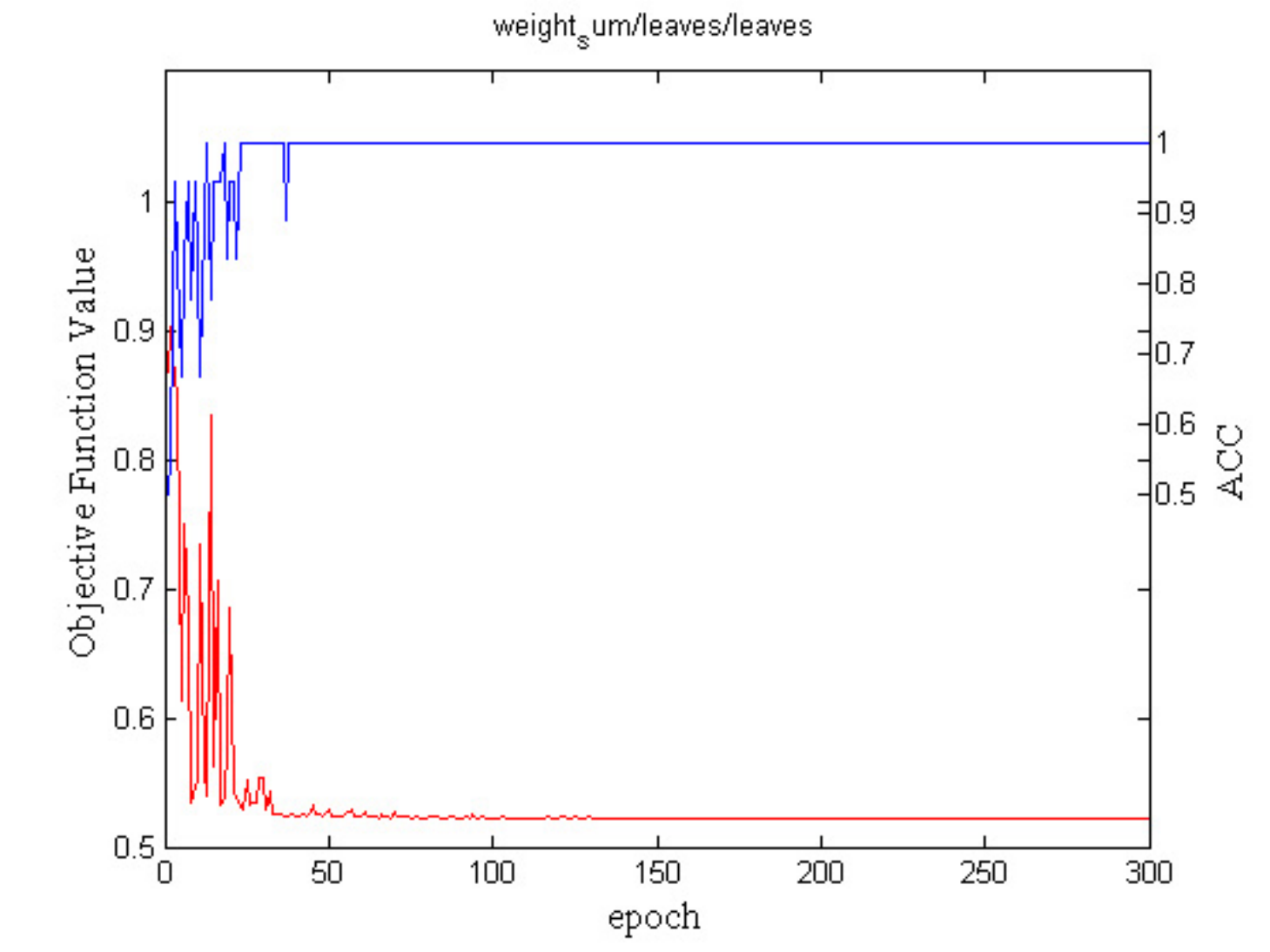}}
\centerline{(a3) leaves-weighted-sum}
\end{minipage}
\qquad
\begin{minipage}{0.30\linewidth}
\centerline{\includegraphics[width=1\textwidth]{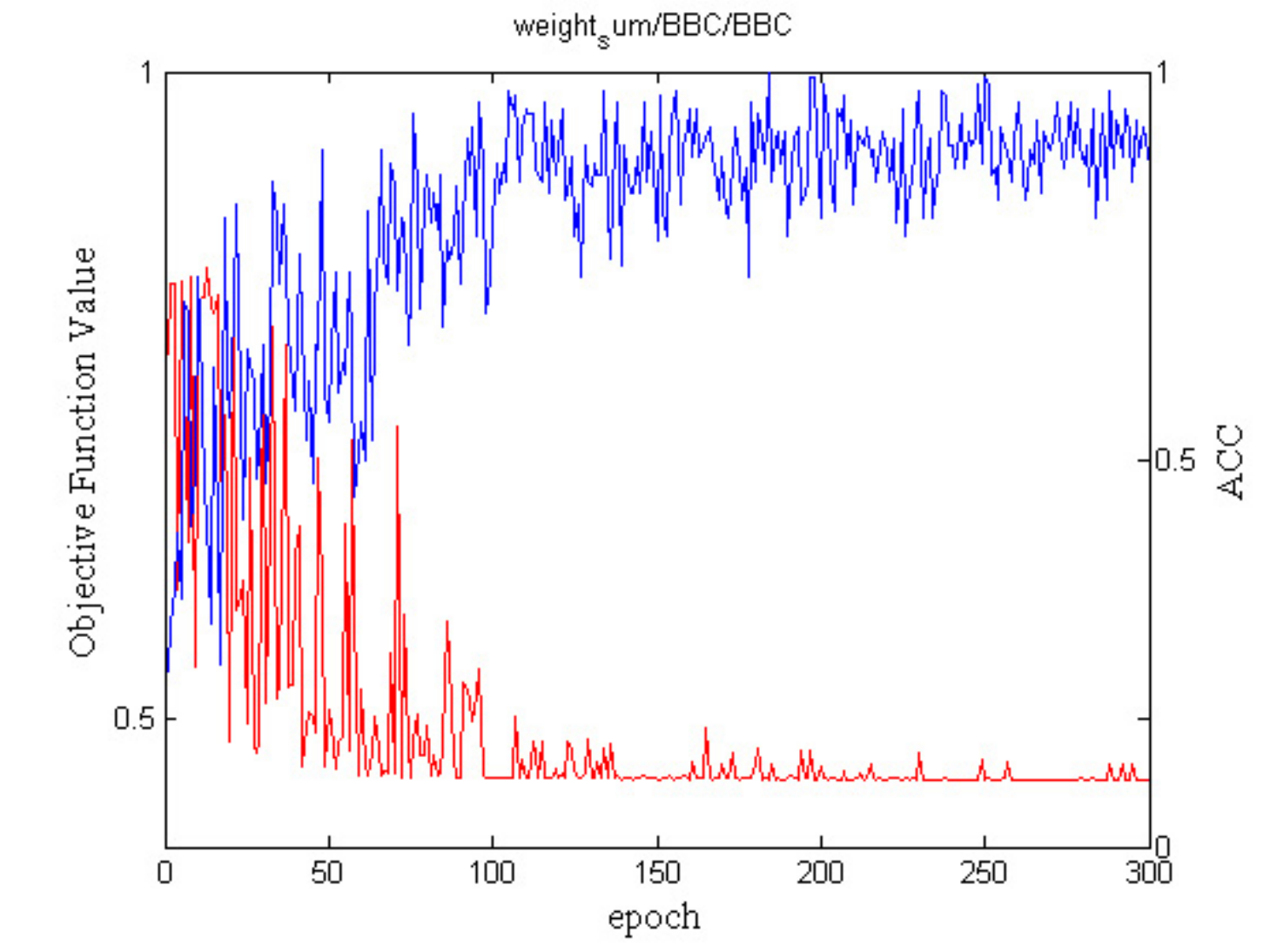}}
\centerline{(b3) BBC-weighted-pooling}
\end{minipage}
\qquad
\begin{minipage}{0.30\linewidth}
\centerline{\includegraphics[width=1\textwidth]{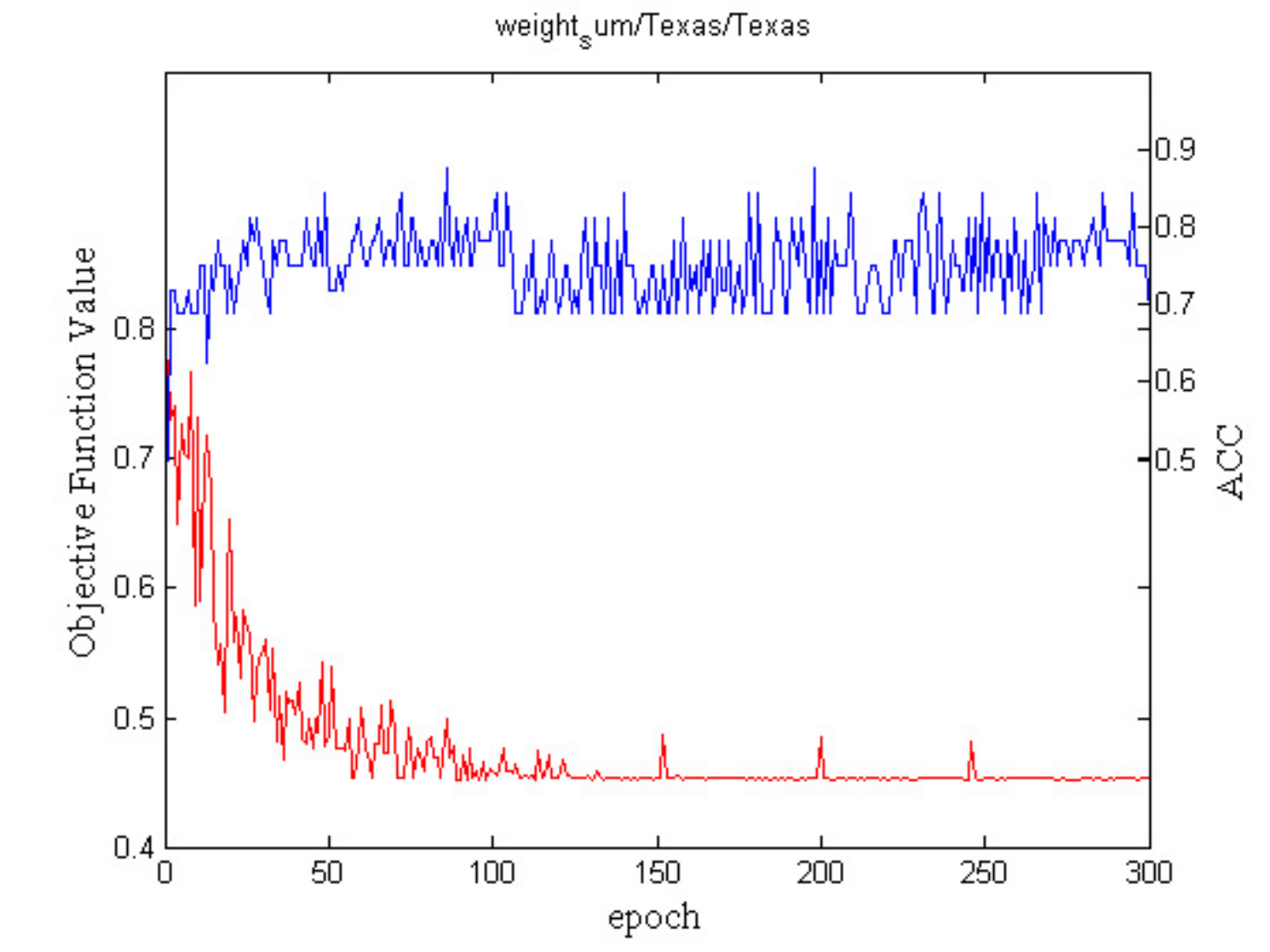}}
\centerline{(c3) Texas-weighted-pooling}
\end{minipage}

\begin{minipage}{0.30\linewidth}
\centerline{\includegraphics[width=1\textwidth]{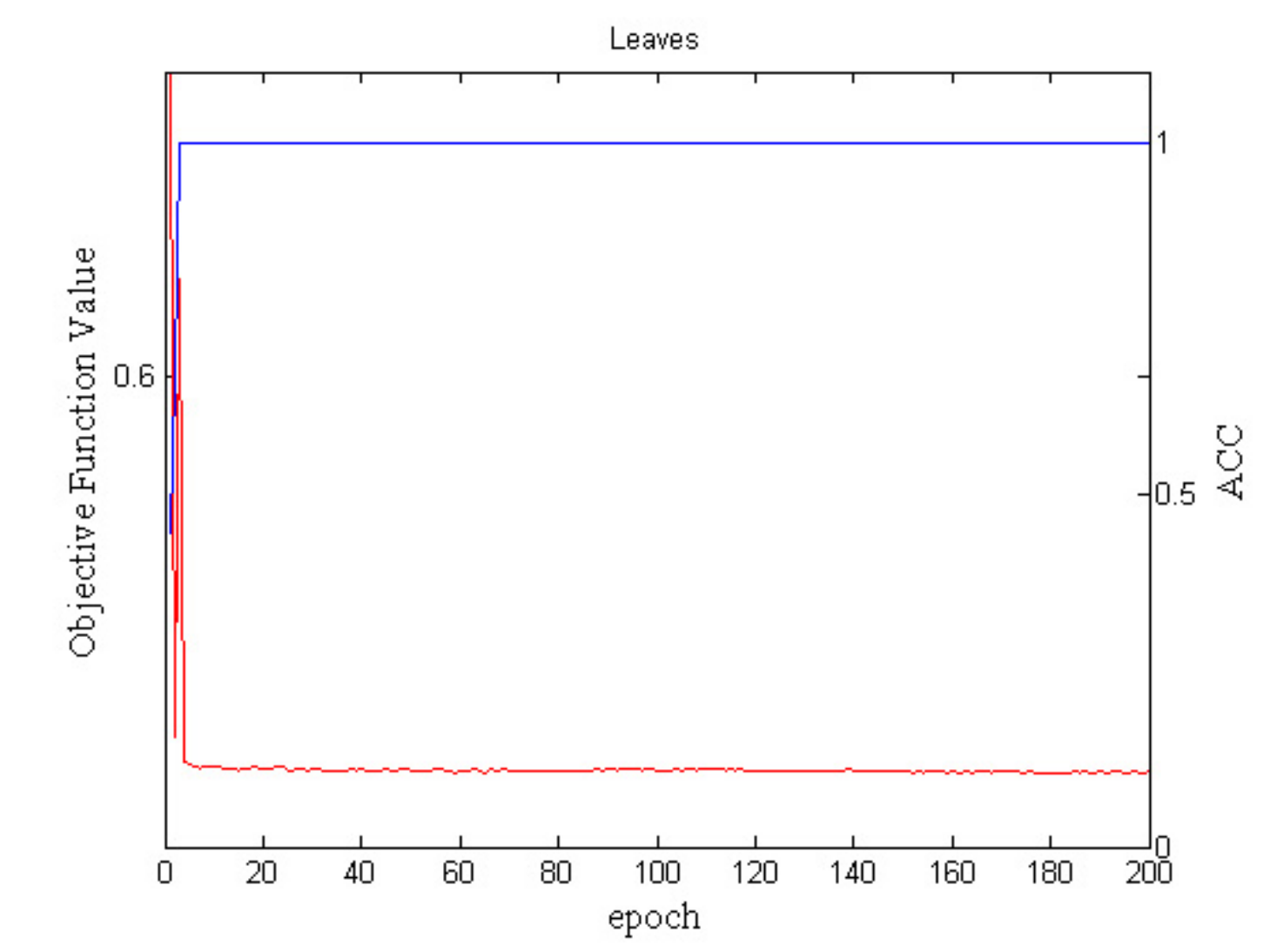}}
\centerline{(a4) leaves-self-attention}
\end{minipage}
\qquad
\begin{minipage}{0.30\linewidth}
\centerline{\includegraphics[width=1\textwidth]{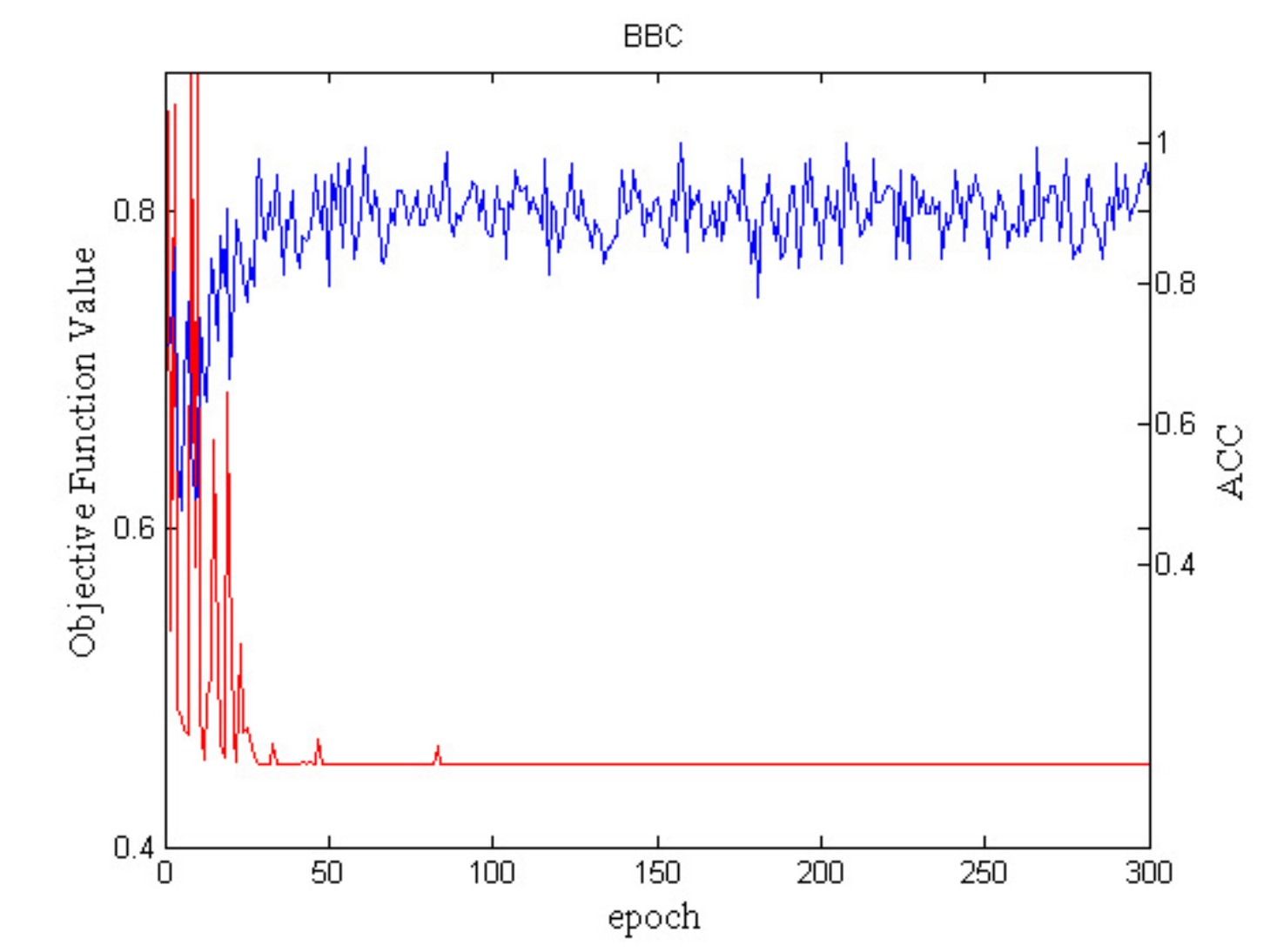}}
\centerline{(b4) BBC-self-attention}
\end{minipage}
\qquad
\begin{minipage}{0.30\linewidth}
\centerline{\includegraphics[width=1\textwidth]{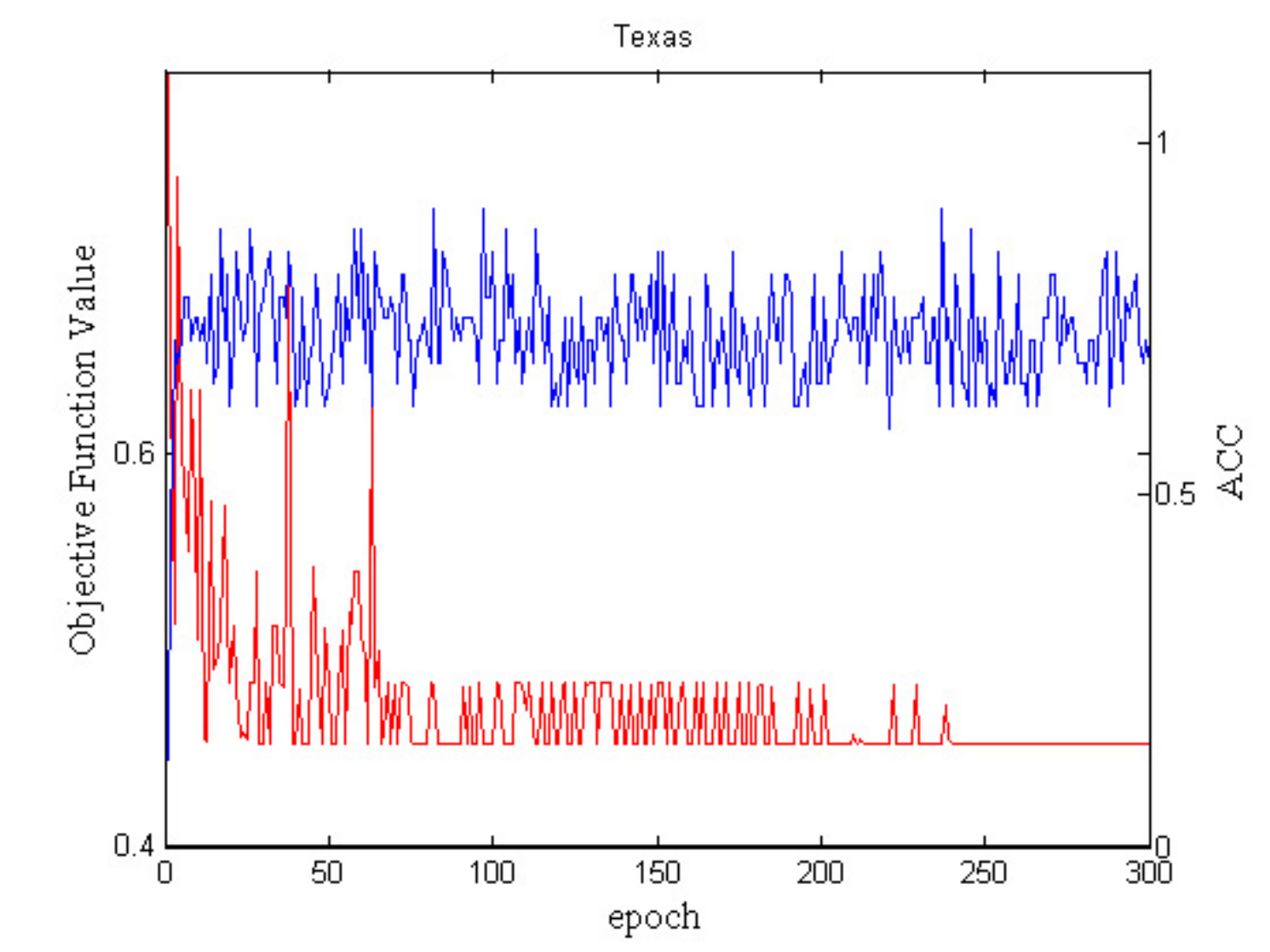}}
\centerline{(c4) Texas-self-attention}
\end{minipage}

\caption{The convergence property of SAMVDP}
\end{center}
\end{figure*}

Due to no publicly available multi-view clustering algorithm based on deep neural network, we generate three baseline models based on deep neural network. These baseline models are exploratory models to validate properties of the self-attention mechanism in SAMVDPC, they separately use max pooling, mean pooling, and weighted summation to fusion all multiple views representations produced by Encoder-Block, find a fusion representation with single complementarity, and then input the fusion representation to fully connected layer to make prediction. And for fair comparison, we use the same settings for baseline models as what we did in SAMVDPC.

\subsection{Configuration and Tricks}

In this subsection, we specify the configuration of SAMVDPC. In Encoder-Block, structures of all encoders are the same, each encoder has one input layer and three hidden layers $l_{1}$, $l_{2}$, and $l_{3}$, the number of number in each hidden layer decrease as the layers of encoder deepens, and the activation function of all hidden layers is ReLU. In self-attention mapping, self-attention MLP has a hidden layer with 300 units  $d\_s$, and we always choose the matrix embedding to have V rows ($d\_c$). In MLP, we use a 2-layer ReLU output MLP with 512 hidden states to output the classification result. For objective function, we usually set $\lambda $ to 0.0001. For the configuration of three baseline models, we use max-pooling, mean-pooling, or weighted summation to take replace of the self-attention mechanism in SAMVDPC, and set $\lambda $ in objective function to 0. The hyper parameters on each data sets are summarized in Table 2.

With regard to the initialization of SAMVDPC weights, In Encoder-Block, we pre-train \emph{V} auto-encoders through minimizing the reconstruction error of each view, and then use the pre-trained parameters of auto-encoders to initialize the corresponding encoder's weight of Encoder-Block. In self-attention MLP and MLP, Xavier is used as the weight initialization method \cite{DBLP:journals/jmlr/GlorotB10}.

In training process, the optimizer algorithm we used is Adam, the learning rate is always initialized to $10^{-3}$, $10^{-4}$, $10^{-5}$, and will decreases gradually with the development of training process. To avoid over fitting, all layers in Encoder-Block and MLP are regularized by dropout regularization in training process, and dropout rate was set to 0.5.

\subsection{Result}

All datasets divided into training, verification and testing data in a ratio of 0.6:0.2:0.2. For SAMVDPC comparison algorithms, and baseline models, we first run each model on each dataset to select hyper parameters that has the best accuracy and generalization performance. And then based on these hyper parameters, we run all algorithms 10 times on each dataset and report the mean values and standard deviation of accuracies.

All the classification results of eight multi-view datasets are summarized in Table 3, and the best result on each dataset is highlighted in boldface. As we can see, the proposed SAMVDPC achieves best accuracy on Texas and YaleFace datasets, and are comparable with other algorithms on the else datasets. The promising result may reason from four aspects: (1) DICS, baseline models, and SAMVDPC are all algorithms exploiting the consistency and complementarity, simultaneously, and compared to GNMF, multiNMF, and MVCC, they all achieve better performance on all datasets; (2) compared to matrix factorization algorithms, the Encoder-Block in both baseline models and SAMVDPC can extract features in a way of effectively fetching consistent information and grasping the underlying common properties of multi-view datasets; (3) compared to baseline models, the complementary with diversity exploited by self-attention mechanism contains more information than max-pooling, mean-pooling, and weighted summation.

\subsection{Convergence Analysis of Training Process}

In order to empirically investigate the convergence property of SAMVDPC, we plot the iterative curves of objective function and the corresponding classification accuracies on three typical data sets, Leaves, BBC, and Texas in Fig. 2. From Fig. 2, we can observe that: (1) the objective function values drop sharply and meanwhile the classification accuracies increase rapidly within the previous rounds of iterative process, and then the objective function and the accuracy curves begin to decrease/grow mildly, finally converge to a value or fluctuate around a constant; (2) with respect to convergence speed, the objective function values of SAMVDPC converge in the least iterations, in contrast, max-pooling corresponds to the most iterations, because max-pooling operation is lossy compression process and the backpropagation process doesn't make full use of information from multiple views data; (3) in respect of convergence result, the objective function of SAMVDPC can finally converge to a fixed value on every dataset, but the objective function of baseline models always finally fluctuate around a constant, what's more, compared to baseline models, we can find that the classification accuracy curves of SAMVDPC often fluctuate within a narrow range. In conclusion, compared to baseline models, SAMVDPC get a better performance on the iterative curves of objective function and the corresponding classification accuracy.

\section{Conclusion and Future Work}

In this paper, we propose a novel multi-view network, called SAMVDPC. The proposed SAMVDPC aims to utilize both complementarity and consistency simultaneously, and give free rein to the potential of deep learning in grasping diversity-promoting complementarity by self-attention mechanism for multi-view representation learning, and then makes prediction based on this representation. In detail, we utilize Encoder-Block to extract feature vectors for each view, leverage self-attention to find complementary information entailing diversity, and make classification by 2-layer MLP. The experimental results on eight real-world data sets also have demonstrated the effectiveness of our proposed paradigm.

For future studies, we plan to change the Encoder-Block into CNN or LSTM to extend our model to complex multi-view data, such as raw image or document. Furthermore, we also plan to investigate the possibility of incorporating meta-learning \cite{bib34} to give better hyper parameters than the human-defined hyper parameters used in this work. Instead of designing routing mechanism according to pre-setting routing strategies, we are also interested in learning to learn not just the learner initialization, but also the learner routing mechanism.


\begin{thebibliography}{0}

\bibitem{bib1} A. Kumar, C. Sminchisescu. 2007. Support kernel machines for object recognition. In 11th ICCV Proceedings, 1-8.
%----------the format for published papers-----------------

\bibitem{bib2} Y. Li, B. Geng, Z.J. Zha, D. Tao, L. Yang, and C. Xu. 2011. Difficulty guided image retrieval using linear multiview embedding. In 19th ACM Multimedia Proceedings, 1169-1172.
%--------- the format for books-------------------------

\bibitem{bib3} C. Wan, R. Pan, and J. Li. 2011. Bi-weighting domain adaptation for cross-language text classification. In 22nd IJCAI Proceedings, 1535-1540.

\bibitem{bib4} B. Xie, Y. Mu, D. Tao, and K. 2011. Huang. m-sne: Multiview stochastic neighbor embedding. IEEE Trans. Systems, Man, and Cybernetics, 41(4):1088-1096.

\bibitem{bib5} Ch. Xu, D. Tao, and C. Xu. 2013. A Survey on Multi-view Learning. arXiv preprint arXiv, 1304.5634.

\bibitem{bib6} K. Chaudhuri, S. M. Kakade, K. Livescu, and K. Sridharan. 2009. Multi-view clustering via canonical correlation analysis. In 26th ICML Proceedings, 129-136.

\bibitem{bib7} J. D. R. Farquhar, D. R. Hardoon, H. Meng, J. Shawe-Taylor, and S. Szedm'k. 2005. Two view learning: SVM-2K, Theory and Practice. In 18th NIPS Proceedings, 355-362.

\bibitem{bib8} D. R. Hardoon, S. Szedm'k, and J. Shawe-Taylor. 2004. Canonical correlation analysis: an overview with application to learning methods. Neural Computation, 16(12):2639-2664.

\bibitem{bib9} M. Kan, S. Shan, H. Zhang, S. Lao, and X Chen. 2016. Multi-view discriminant analysis. IEEE Trans. Pattern Anal. Mach. Intell, 38(1):188-194.

\bibitem{bib10} A. Sharma, A. Kumar, H. Daum', D. and W. Jacobs. 2012. Generalized multiview analysis: a discriminative latent space. In 20th CVPR Proceedings, 2160-2167.

\bibitem{bib11} Z. Guan, L. Zhang, J. Peng, and J Fan. 2015. Multi-view concept learning for data representation. IEEE Trans. Knowl. Data Eng., 27(11): 3016-3028.

\bibitem{bib12} J. Gao, J. Han, J. Liu, and C. Wang. 2013. Multi-view clustering via joint nonnegative matrix factorization. In 13th SDM Proceedings, 252-260.

\bibitem{bib13} H. Wang, Y. Yang, and T. Li. 2016. Multi-view clustering via concept factorization with local manifold regularization. In 16th ICDM Proceedings, 1245-1250

\bibitem{bib14} A. Blum, T. M. Mitchell. 1998. Combining labeled and unlabeled data with co-training. In 16th COLT Proceedings, 92-100.

\bibitem{bib15} A. Kumar, and H. Daum'. 2011. A co-training approach for multi-view spectral clustering. In 28th ICML Proceedings, 393-400.

\bibitem{bib16} W. Wang, Z. Zhou. 2010. A new analysis of co-training. In 27th ICML Proceedings, 1135-1142.

\bibitem{bib17} M. Zhang, Z. Zhou. 2011. CoTrade: confident co-training with data editing. IEEE Trans. Systems, Man, and Cybernetics, 41(6):1612-1626.

\bibitem{bib18} Z. Zhang, Z. Qin, P. Li, Q. Yang, and J. Shao. 2018. Multi-view discriminative learning via joint non-negative matrix factorization. In 23rd DASFAA Proceedings, 542-557.

\bibitem{bib19} A. P. Singh, G. J. Gordon. 2008. Relational learning via collective matrix factorization. In 14th KDD Proceedings, 650-658.

\bibitem{bib20} N. Srivastava, R. Salakhutdinov. 2014. Multimodal learning with deep Boltzmann machines. Journal of Machine Learning Research, 15(1):2949-2980.

\bibitem{bib21} J. Ngiam, A. Khosla, M. Kim, J. Nam, H. Lee, and A. Y. Ng: 2011. Multimodal Deep Learning. In 1rth KDD Proceedings, 689-696.

\bibitem{bib22} K. Cho, B. V. Merrienboer, ?. G'l?ehre, D. Bahdanau, F. Bougares, H. Schwenk, and Y. Bengio. 2014. Learning Phrase Representations using RNN Encoder-Decoder for Statistical Machine Translation. In 2014 EMNLP Proceedings, 1724-1734.

\bibitem{bib23} H. Su, S. Maji, E. Kalogerakis, and E. G. Learned-Miller. 2015. Multi-view Convolutional Neural Networks for 3D Shape Recognition. In 2015 ICCV Proceedings, 945-953.

\bibitem{bib24} D. Bahdanau, K. Cho, and Y. Bengio. 2015. Neural Machine Translation by Jointly Learning to Align and Translate. In 3rd ICLR Proceedings.

\bibitem{bib25} I. Sutskever, O. Vinyals, and Q. V. Le. 2014. Sequence to Sequence Learning with Neural Networks. In 28t hNIPS Proceedings, 3104-3112.

\bibitem{bib26} K. Xu, J. Ba, R. Kiros, K. Cho, A. C. Courville, R. Salakhutdinov, R. S. Zemel, and Y. Bengio. 2015. Show, Attend and Tell: Neural Image Caption Generation with Visual Attention. In 32nd ICML Proceedings, 2048-2057.

\bibitem{bib27} A. M. Rush, S. Chopra, and J. Weston. 2015. A Neural Attention Model for Abstractive Sentence Summarization. In 2015 EMNLP Proceedings, 379-389.

\bibitem{bib28} J. Cheng, L. Dong, and M. Lapata. 2015. Long Short-Term Memory-Networks for Machine Reading. In 2016 EMNLP Proceedings, 551-561.

\bibitem{bib29} A. P. Parikh, O. T?ckstr?m, D. Das, and J. Uszkoreit. 2016. A Decomposable Attention Model for Natural Language Inference. In 2016 EMNLP Proceedings, 2249-2255.

\bibitem{bib30} Z. Lin, M. Feng, C. N. d. Santos, M. Yu, B. Xiang, B. Zhou, and Y. Bengio. 2017. A Structured Self-Attentive Sentence Embedding. In 5th ICLR Proceedings.

\bibitem{bib31} A. Vaswani, N. Shazeer, N. Parmar, J. Uszkoreit, L. Jones, A. N. Gomez, L. Kaiser, and I. Polosukhin. 2017. Attention is All you Need. In 31t h NIPS Proceedings, 6000-6010.

\bibitem{bib32} Z. Wang, X. Kong, H. Fu, M. Li, and Y. Zhang. 2015. Feature extraction via multi-view non-negative matrix factorization with local graph regularization. In 2015 ICIP Proceedings, 3500-3504.

\bibitem{bib33} X. Glorot, Y. Bengio: 2010. Understanding the difficulty of training deep feedforward neural networks. In 13th AISTATS Proceedings, 249-256.

\bibitem{bib34} C. Finn, P. Abbeel, and S. Levine. 2017. Model-agnostic meta-learning for fast adaptation of deep networks. In 34th ICML Proceedings, 1126-1135.

\end{thebibliography}
\end{document}